\documentclass{article}
\usepackage{arxiv}
\usepackage{amsmath} 
\usepackage{amssymb}
\usepackage{mathtools}
\usepackage{siunitx}
\usepackage{mathtools}
\usepackage{algorithm}
\usepackage{algpseudocode}
\usepackage{listings}
\usepackage{xcolor}
\usepackage{hyperref}

\DeclarePairedDelimiterXPP\BigOSI[2]%
  {\mathcal{O}}{(}{)}{}%
  {\SI{#1}{#2}}

\usepackage[utf8]{inputenc} 
\usepackage[T1]{fontenc}    
\usepackage{hyperref}       
\usepackage{url}            
\usepackage{booktabs}       
\usepackage{amsfonts}       
\usepackage{nicefrac}       
\usepackage{microtype}      
\usepackage{lipsum}
\usepackage{graphicx}

\title{Cottention: Linear Transformers With Cosine Attention}

\author{
 Gabriel Mongaras \\
  Lyle School of Engineering\\
  Southern Methodist University\\
  Dallas, TX 75205 \\
  \texttt{gabriel@mongaras.com} \\
  \And
 Trevor Dohm \\
  Lyle School of Engineering\\
  Southern Methodist University\\
  Dallas, TX 75205 \\
  \texttt{trevordohm@gmail.com} \\
  \And
 Eric Larson \\
  Lyle School of Engineering\\
  Southern Methodist University\\
  Dallas, TX 75205 \\
  \texttt{eclarson@smu.edu} \\
}

\begin{document}
\maketitle
\begin{abstract}
Attention mechanisms, particularly softmax attention, have been instrumental in the success of transformer-based models such as GPT. However, the quadratic memory complexity of softmax attention with respect to sequence length poses significant challenges for processing longer sequences. We introduce Cottention, a novel attention mechanism that replaces the softmax operation with cosine similarity. By leveraging the properties of cosine similarity and rearranging the attention equation, Cottention achieves native linear memory complexity with respect to sequence length, making it inherently more memory-efficient than softmax attention. We demonstrate that Cottention can be reformulated as a recurrent neural network (RNN) with a finite hidden state, allowing for constant memory usage during inference. We evaluate Cottention on both the bidirectional BERT and causal GPT tasks, demonstrating comparable performance to softmax attention while significantly reducing memory requirements. To ensure efficient computation, we develop a custom CUDA kernel for Cottention. Our results show that Cottention is a promising alternative to softmax attention, enabling the processing of longer sequences without sacrificing performance, due to its native linear memory complexity and ability to maintain a constant memory footprint during inference.\footnote{Code: \url{https://github.com/gmongaras/Cottention_Transformer}}
\end{abstract}

\section{Introduction}

Transformer models have achieved unprecedented success in various applications ranging from natural language processing to computer vision \cite{vaswani2017attention, devlin2018bert, brown2020language, dosovitskiy2020image}. Central to these model's capability is the attention mechanism, a powerful computation that allows models to adapt representations based on focused context of the entire sequence \cite{bahdanau2014neural, luong2015effective}. However, the attention mechanism's expressiveness comes at the cost of computation as the sequence length increases, due to the quadratic complexity in both time and memory of the softmax operation \cite{tay2022efficient, zaheer2020big}. This limitation has spurred interest in developing more efficient attention mechanisms that can handle longer sequences without such a steep cost.

Several attempts have been made to address this issue, including the introduction of sub-quadratic time architectures like linear attention \cite{katharopoulos2020transformers, choromanski2020rethinking, peng2021random}, gated convolutional and recurrent models \cite{dauphin2017language, lei2021attention}, and structured state space models (SSMs) \cite{gu2021efficiently, gu2021combining, gupta2022diagonal}. While these efforts have yielded improvements in computational efficiency, they often fall short in matching the performance of traditional attention mechanisms on key tasks, particularly in the domain of language processing. A critical examination of these models reveals that their primary limitation could be diminished capacity for content-based reasoning \cite{tay2022efficient}, although additional investigation is needed to fully examine the performance gap.

Recent works have explored the potential of alternative similarity measures to replace the softmax operation in attention mechanisms \cite{kitaev2020reformer, choromanski2020rethinking, qin2022cosformer}. These approaches have shown promising results in reducing computational complexity while maintaining competitive performance. Other works have also investigated the use of cosine similarity in various contexts \cite{rabe2021self, nguyen2023enhancing, luo2018cosine, crowson2024scalable, liu2022swin}. However, for each of these, their application has been limited to specific domains or doesn't explore optimal stabilization techniques.

In this work, we propose a novel attention mechanism, Cottention (Cosine Attention), which leverages the properties of cosine similarity to achieve linear complexity concerning sequence length. We generalize cosine attention to work with arbitrary-length sequences and apply it to the text domain, demonstrating its effectiveness on a range of language tasks. Our approach addresses the stability issues encountered in previous works without the need for additional constraints or modifications \cite{wang2020linformer}. Additionally, we show that although cosine attention is linear with respect to sequence length, it retains similar accuracy to softmax attention.

\section{Related Work}

\subsection{Softmax Attention}
\label{sec:softmax_attention}

Softmax attention, as formalized in equation \ref{EQ:softmax}, has been the standard in transformer models since their introduction \cite{vaswani2017attention}. The attention mechanism computes a weighted sum of the value vectors, where the weights are obtained by applying a softmax function to the scaled dot-product of the query and key vectors. In the multi-head setting, the embedding dimension is split into $ H $ heads, and the attention is applied independently for each head before recombining and projecting the outputs \cite{devlin2018bert, radford2018improving}. We provide this formulation below.

Let $ Q $, $ K $, and $ V $ denote the queries, keys, and values obtained by projecting the input $ x $, where $ N $ is the batch size, $ H $ is the number of heads, $ s $ is the sequence length, $ d_{key} $ is the inner attention dimension, and $ d_{model} $ is the model dimension. The input is a batch of sequences $ x \in \mathbb{R}^{N \times s \times \text{d}_\text{model}} $ and the projection matrices are $ W_Q \in \mathbb{R}^{\text{d}_\text{model} \times \text{d}_\text{key}} $, $ W_K \in \mathbb{R}^{\text{d}_\text{model} \times \text{d}_\text{key}} $, and $ W_V \in \mathbb{R}^{\text{d}_\text{model} \times \text{d}_\text{value}} $. After projection, we convert this into multihead attention by splitting the queries, keys, and values along the dimension index with $d_\text{H\_key} = d_\text{key} / H$ being the per-head key dimension and $d_\text{H\_value} = d_\text{value} / H$ being the per-head model dimension.
\vspace{4px}
\begin{equation*}
Q = x W_Q \in \mathbb{R}^{N \times H \times s \times \text{d}_\text{H\_key}} \ \ \ \ \ \ K = x W_K \in \mathbb{R}^{N \times H \times s \times \text{d}_\text{H\_key}} \ \ \ \ \ \ V = x W_V \in \mathbb{R}^{N \times H \times s \times \text{d}_\text{H\_value}}
\end{equation*}
\begin{equation}
\text{Attention}(Q, K, V) = \text{softmax} \left( \frac{Q K^T}{\sqrt{d_k}} \right) V
\label{EQ:softmax}
\end{equation}

The softmax attention mechanism has a quadratic complexity in both time and memory, as the computation of $ QK^T $ is quadratic in the sequence length $s$ and the dimensionality $ d $ of each token. Specifically, the operation has a time complexity of $ \BigOSI{}{s^2 d} $ and a maximum memory usage of $ \BigOSI{}{s^2} $. While this quadratic complexity is manageable for short sequences, the recent success of large language models (LLMs) \cite{brown2020language, chowdhery2023palm, hoffmann2022training} has necessitated the processing of longer sequences for in-context learning, rendering the quadratic complexity intractable \cite{tay2022efficient}.




\subsection{Subquadratic Attention}

To address the computational challenges posed by softmax attention, various subquadratic attention mechanisms have been proposed. These approaches aim to reduce the time and memory complexity while maintaining the expressiveness and performance of the attention mechanism. We mention some notable approaches in this section.

Sparse attention mechanisms \cite{child2019generating, kitaev2020reformer, beltagy2020longformer, zaheer2020big} aim to reduce the computational complexity by attending to only a subset of the input sequence. These methods employ techniques such as local attention, global attention, and strided attention to capture both short-range and long-range dependencies efficiently. By reducing the number of attended positions, sparse attention mechanisms can achieve subquadratic complexity, making them suitable for processing longer sequences. However, the selection of the attended positions is crucial for maintaining model performance, and these methods may require additional tuning and domain knowledge.

Linear attention methods \cite{katharopoulos2020transformers, choromanski2020rethinking, peng2021random} approximate the softmax attention by expressing it as a linear combination of kernel functions. By leveraging the associative property of matrix multiplication, these methods achieve a linear complexity with respect to the sequence length. However, the linear approximation may not fully capture the expressive power of the softmax attention, leading to potential performance degradation \cite{tay2022efficient}.

Gated convolutional and recurrent models \cite{dauphin2017language, lei2021attention} incorporate gating mechanisms to control the flow of information in the network. These models can process sequences with a linear complexity, but, for convolutional models, their receptive field is typically limited to a fixed context window. Consequently, they may struggle to capture long-range dependencies that are crucial for many language tasks \cite{tay2022efficient}.

Structured state space models (SSMs) \cite{gu2021efficiently, gu2021combining, gupta2022diagonal} have emerged as a promising alternative to attention mechanisms. SSMs model the input sequence as a continuous-time process and leverage the properties of state space representations to achieve efficient computation. By parameterizing the state transitions with structured matrices, such as diagonal or low-rank matrices, SSMs can model long-range dependencies with a linear complexity. However, the performance of SSMs on language tasks has been limited compared to softmax attention \cite{gupta2022diagonal}.

Flash Attention \cite{dao2022flashattention} is an algorithm that optimizes the softmax attention operation by exploiting the sparsity and structure of the attention matrix. It achieves significant speedups and memory savings compared to the standard softmax attention, enabling the training of larger models with longer sequences. However, Flash Attention is not natively linear and is a reformulation of quadratic softmax attention. In contrast, our method is natively linear, with an ease of implementation and has the associativity matrix property, which may allow for certain matrix properties to be leveraged for future work. We discuss the implications of this property and its potential for future research directions in section \ref{sec:future_work}.

Recent works have further improved upon Flash Attention by introducing novel attention mechanisms that scale to even longer sequences. Ring Attention \cite{liu2023ring} proposes a blockwise parallel transformer architecture that enables near-infinite context by efficiently propagating information across blocks. Striped Attention \cite{brandon2023striped} further optimizes Ring Attention by introducing a striped attention pattern that reduces the computational complexity and memory usage. These advancements have enabled the training of models on million-length videos and language sequences \cite{liu2024world}.

Previous works have explored alternatives to the softmax function in attention mechanisms. The Reformer model \cite{kitaev2020reformer} employs locality-sensitive hashing (LSH) to approximate similarity between queries and keys, while the Performer model \cite{choromanski2020rethinking} approximates the softmax attention using random feature maps, reducing computational complexity. The Cosformer model \cite{qin2022cosformer} implements a cosine-based distance re-weighting mechanism in softmax attention and applies normalization techniques to stabilize training. However, these approaches have limitations that hinder their widespread adoption. The Reformer model's reliance on LSH may not be suitable for all sequence lengths and introduces additional computational overhead. The Performer model has been primarily evaluated on language tasks and may require further investigation for its applicability to other domains. Similarly, the Cosformer model has been tested on a limited range of tasks and may need additional experiments to validate its effectiveness and stability across various scenarios. 

Other works have also investigated the use of cosine similarity in various contexts. \cite{rabe2021self} employed cosine similarity in self-supervised learning for visual representations, while \cite{nguyen2023enhancing} utilized it in enhancing cross-modal retrieval. \cite{luo2018cosine} and \cite{crowson2024scalable} explored cosine similarity in the context of convolutional neural networks and scalable attention mechanisms, respectively. However, these approaches either encountered stability issues or reverted to using the standard softmax attention, suggesting room for further improvement in leveraging cosine similarity effectively in attention mechanisms.

In contrast, our proposed cosine attention mechanism generalizes cosine attention to work with arbitrary-length sequences and applies it to the text domain. We address the stability issues encountered in previous works without the need for additional constraints or modifications, making Cottention a drop-in replacement for softmax attention in both bidirectional and causal transformer models. By combining the benefits of cosine similarity with the advancements in subquadratic attention mechanisms, Cottention has the potential to further push the boundaries of efficient and scalable attention-based models.


\section{Algorithm}
\subsection{Cosine Similarity}
Cosine similarity is a measure of similarity between two non-zero vectors in an inner product space, defined as the cosine of the angle between them. This measure is intrinsically bounded within the interval $ [-1, 1] $, where a value of $ 1 $ indicates that the vectors are pointing in the same direction (i.e., they are positively correlated), a value of $ -1 $ indicates that the vectors are pointing in opposite directions (i.e., they are negatively correlated), and a value of $ 0 $ indicates that the vectors are orthogonal (i.e., they are uncorrelated). Cosine similarity is given by equation \ref{EQ:cosine_sim}
\vspace{2px}
\begin{equation}
    \text{Sim}(X, Y) = \frac{X \cdot Y^T}{ \lVert X \rVert_2 \lVert Y \rVert_2 } = \frac{X}{ \lVert X \rVert_2 } \cdot \frac{Y^T}{ \lVert Y \rVert_2 }
\label{EQ:cosine_sim}
\end{equation}

Cosine similarity is widely used in various fields, including machine learning, data mining, and information retrieval, due to its computational efficiency and effectiveness in measuring the normalized similarity between high-dimensional vectors. The use of the dot product operation in the numerator of equation \ref{EQ:cosine_sim} makes it particularly efficient to compute on GPU hardware, which is optimized for parallel vector operations.

\subsection{Cosine Attention}
Cosine attention leverages the property that the norm of the input matrices can be decoupled, as in the righthand side of \eqref{EQ:cosine_sim}, by first L2 normalizing each matrix along row vectors, then performing the matrix multiplication between the normalized $ Q $ and $ K^T $:
\vspace{6px}
\begin{equation}
\text{CosAttention}(Q, K, V) = \text{Sim}(Q, K) \cdot V = \mathcal{N}(Q) \cdot \mathcal{N}(K)^T \cdot V
\label{EQ:cosine}
\end{equation}

In the above equation, $ \mathcal{N}(X) = \frac{X}{\lVert X \rVert_2} $ represents the L2 normalization operation applied to each row vector of the input matrix.
By replacing the softmax operation in the standard attention mechanism with the cosine similarity function defined in \eqref{EQ:cosine_sim}, we obtain the cosine attention formulation presented in \eqref{EQ:cosine}. This formulation allows for efficient computation of the attention weights by decoupling the normalization step from the matrix multiplication, enabling the use of optimized linear algebra routines, which we implement in CUDA. 


\subsection{Stabilizing Cosine Attention}

Unlike softmax attention \eqref{EQ:softmax}, cosine attention \eqref{EQ:cosine} can be unstable during training. The instability arises from the similarity matrix $ \text{Sim}(Q, K) $, which has a maximum row-wise sum of $ s $, while the softmax attention matrix $ \text{softmax}(Q K^T) $ has a row-wise sum of 1. The large magnitude of the similarity matrix can lead to unstable training. To address this issue, we propose a stabilized formulation of cosine attention, given as:
\vspace{5px}
\begin{equation}
\text{CosAttention}(Q, K, V) = \frac{1}{s^{\sigma(m)}} \odot \text{Sim}(Q, K) \cdot V
\label{EQ:cosine2}
\end{equation}

Dividing the similarity matrix by the sequence length $ s $ ensures that the maximum row-wise sum of the similarity matrix falls within the range $ [0, 1] $. However, we find that the row-wise sums tend to be much smaller than the sequence length upon initialization, allowing for a relaxation of this restriction. 

To introduce this flexibility, we divide the value by the sequence length raised to the power of a learned scalar $ m $, which is passed through a sigmoid function. This allows the model to learn to divide by a minimum value of $ 1 $ or a maximum value of $ s $. The scalar $ m $ is initialized to $ 0.5 $ for each attention head. The division operation is applied to the value matrix before computing the attention output, but it could equivalently be applied during or after the attention computation. 

The introduction of the learned scalar $ m $ for each attention head adds a small number of parameters to the model, equal to $ (\text{num\_heads}) * (\text{num\_layers}) $. In the case of our GPT model with 20 layers and 16 heads, the parameter count amounts to an additional 320 parameters. While we found the initialization of $ m = 0.5 $ to work well, there may be room for improvement in the initialization strategy and/or the choice of $ m $ to further stabilize model training. We leave the exploration of optimal initialization strategies to future work.

\section{Subquadratic Cosine Attention}

Removing the dimension subscripts for brevity, cosine attention can achieve $d^2$ memory complexity in the bidirectional case by rearranging the computation order:
\vspace{7px}
\begin{equation}
\left[ \mathcal{N}(Q) \cdot\mathcal{N}(K)^T \right]\cdot V = \mathcal{N}(Q)\cdot \left[ \mathcal{N}(K)^T \cdot V \right]
\label{EQ:subQuadCos}
\end{equation}

The left side of the equation computes the inner product $ Q K^T $ matrix first, which requires $ s^2 $ memory and is optimal when $ s < d_{H\_key} $. The right side of the equation computes the outer product $ K^T V $ first, which requires $ d^2 $ memory and is optimal when $ s > d_{H\_key} $. A bidirectional model such as BERT can utilize the right side of equation \ref{EQ:subQuadCos}. When a padding mask is needed, it can be applied to the input $ Q $, $ K $, and $ V $ matrices before performing the attention operation.

However, in the causal case with an attention mask, the right-hand side of \eqref{EQ:subQuadCos} is not valid, and achieving $ d^2 $ memory complexity is more challenging. To analyze this problem, we can simplify the attention operation \eqref{EQ:attn_1} to be between three matrices, with operations like normalization performed before or after the attention operation, where $ \odot $ denotes the Hadamard product and $ M $ denotes the applied mask:
\vspace{5px}
\begin{equation}
    O = (Q \cdot K^T \odot M) \cdot V
\label{EQ:attn_1}
\end{equation}

Softmax attention uses an additive upper triangle matrix of $ -\infty $ and $ 0 $ as the mask, leveraging the softmax operation to turn $ -\infty $ values into zeros. Cosine attention, on the other hand, uses a binary multiplicative mask with the lower triangular populated with ones and zeros above the diagonal. Due to this mask, rearranging the equation cannot yield a $ d^2 $ operation using dense operations, as in the bidirectional case. However, the equation can be rearranged to be $ d^2 $ using altered groupings of the matrix operations. Rather, we perform strategic cumulative and reductive sums to maintain the previous structure given in equation \ref{EQ:attn_1}. The forward pass can be rewritten as in equation \ref{EQ:attn_2}:
\vspace{5px}
\begin{equation*}
\bar{Q} \in \mathbb{R}^{N \times H \times s \times 1 \times \text{d}_\text{H\_key}} \ \ \ \ \ \ \bar{K} \in \mathbb{R}^{N \times H \times s \times 1 \times \text{d}_\text{H\_key}} \ \ \ \ \ \ V \in \mathbb{R}^{N \times H \times s \times \text{d}_\text{H\_value} \times 1}
\end{equation*}
\vspace{-4px}
\begin{equation}
O = \left( \left( \bar{V} \odot \bar{K} \right).\text{cumsum}(-3) \odot \bar{Q} \right).\text{sum}(-1)
\label{EQ:attn_2}
\end{equation}

Here, the $ \bar{X}$ denotes a shift in the dimensionality to broadcast operations correctly. In tensor processing libraries, such as PyTorch, this is defined as an `unsqueeze' method. While this operation can be coded easily, the naive implementation results in $ sd^2 $ memory usage, which is much worse than the $ s^2 $ memory usage required to store the attention matrix for any reasonable head dimension size. To make this operation efficient, a custom CUDA kernel can be utilized, allowing the forward pass to be computed with a single kernel, as shown in our codebase \footnotemark.

The kernel strategy operates on the intermediate $d^2$ block, where the inner index corresponds to the key dimension and the outer index corresponds to the value dimension. This approach differs from the standard matrix multiplication strategy, which typically involves tiling the output matrix. In this kernel, each thread is responsible for computing a portion of the key dimension, while each block computes an entire row for all segments of the sequence. This design allows for efficient parallel computation, as threads within a block can work collaboratively on different parts of the key dimension, and multiple blocks can process different rows simultaneously. By leveraging this strategy, the kernel can effectively utilize the available computational resources and optimize the performance of the cosine attention operation.


Since an operation written in native CUDA does not have autograd, the backward pass must be manually calculated for all input tensors. The gradients with respect to each input are given by:
\vspace{4px}
\begin{equation*}
\begin{aligned}
\frac{\partial L}{\partial Q} = \left( \frac{\partial L}{\partial O} V^T \odot M \right) K, \qquad
\frac{\partial L}{\partial K} = Q^T \left( \frac{\partial L}{\partial O} V^T \odot M \right) ^ T \qquad
\frac{\partial L}{\partial V} = \left( Q K^T \odot M \right) ^T \frac{\partial L}{\partial O}
\end{aligned}
\end{equation*}

The gradient of $ Q $ is simply another forward pass with the gradient of the output, $ Q $, and $ V $ instead of $Q$, $ K $, and $ V $. The gradients of $ K $ and $ V $ can also be computed by another forward pass, but the cumulative sum goes backward in time, accumulating values from the end of the sequence. A naive implementation of cosine attention using PyTorch operations can be found in Appendix \ref{App_Code}.

\section{RNN Reformulation}


Cosine attention can be reformulated as a recurrent neural network (RNN) by interpreting the cumulative sum operation in the forward pass \eqref{EQ:attn_2} as a recurrent computation. A similar analysis was conducted by \cite{katharopoulos2020transformers}, which also revealed that decoupling the softmax operation allows for reformulating the attention mechanism as a recurrent operation. This reformulation reveals interesting properties and provides a fresh perspective on the inner workings of cosine attention. We consider:
\vspace{6px}
\begin{equation}
H_t = H_{t-1} + (\bar{V} \odot \bar{K}) ,
\qquad
O_t = (H_t \odot \bar{Q})\text{.sum}(-1)
\label{EQ:hidden_state_obs}
\end{equation}

The forward pass in \eqref{EQ:attn_2} can be translated into an equivalent RNN formulation, where the hidden state is computed according to equation \ref{EQ:hidden_state_obs} (left). At each timestep $ t $, the hidden state stores the keys and values from all previous tokens. It is updated by performing an element-wise outer product between the keys and values at timestep $ t $ and accumulating the result with the hidden state from the preceding timestep. The hidden state matrix has dimensions $ (N, H, d_{H\_model}, d_{H\_key}) $, which can be interpreted as $ H $ parallel RNNs operating independently, each with its own state that encapsulates the keys and values from all timesteps prior to $ t $. This parallel processing architecture enables efficient computation and allows the model to capture diverse aspects of the input sequence concurrently.


To generate the output of the attention mechanism at timestep $ t $, the hidden state is observed through equation \ref{EQ:hidden_state_obs} (right). This equation computes an inner product between the query and each row vector of the hidden state matrix, yielding a matrix of shape $ (N, H, d_{H\_model}) $ that represents the attended output at timestep $ t $. By performing this inner product, the model effectively assesses the relevance of each hidden state element to the current query, enabling it to focus on the most salient information for generating the output.



\begin{figure*}[t]
\centering
\begin{minipage}{0.45\linewidth}
\centering
\includegraphics[width=0.80\linewidth]{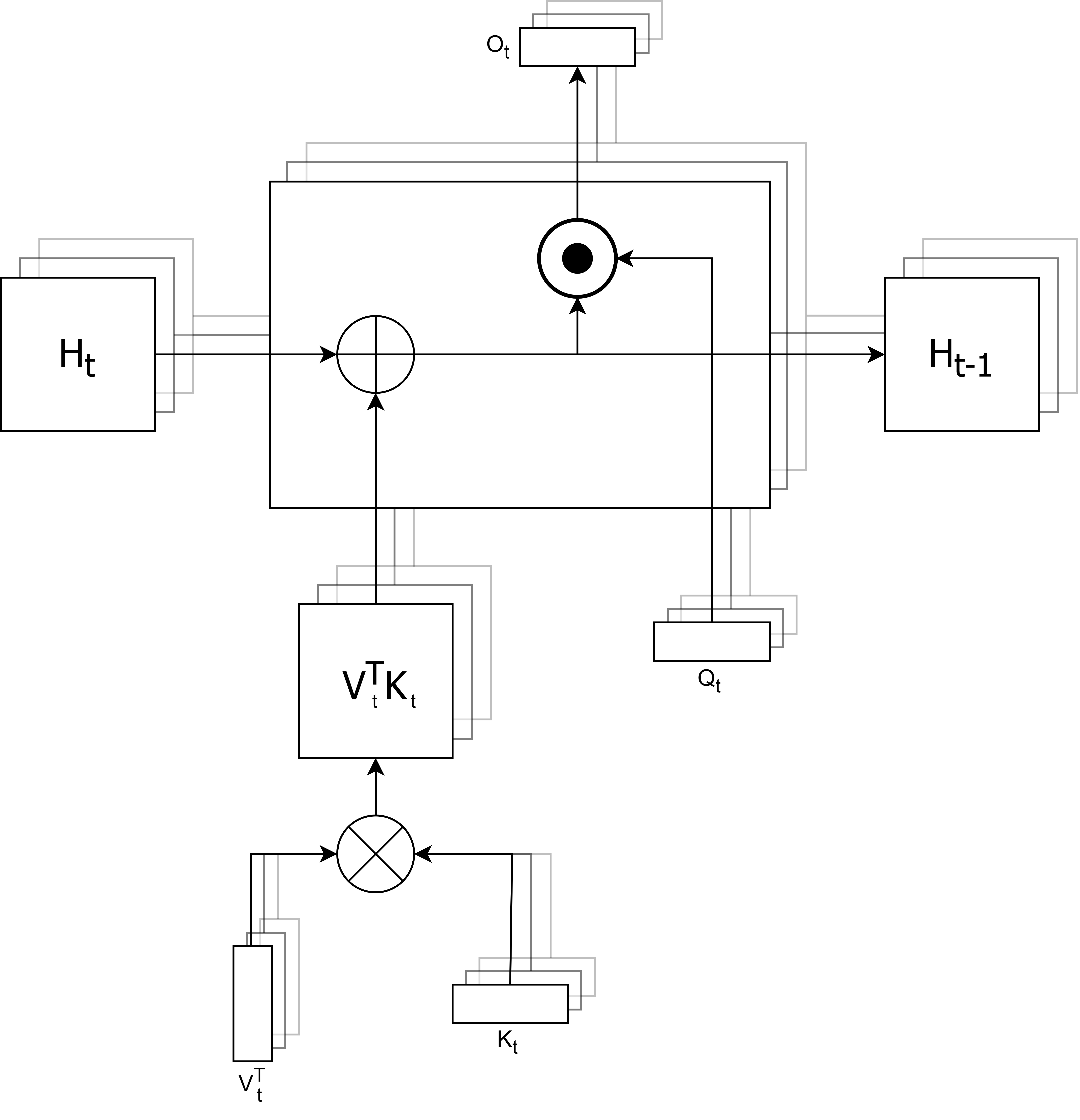}
\caption{Recurrent neural network representation of cosine attention where the queries, keys, and values are of shape $ (N, H, (d_{H\_key/H\_key/H\_value})) $ and the hidden state is shape $ (N, H, d_{H\_value}, d_{H\_key}) $. \\ $ \otimes $ represents an outer product, $ \odot $ represents an inner product, and $ \oplus $ is a position-wise addition. The hidden state $ H_0 $ is initialized to the zero matrix or null matrix.}
\end{minipage}
\hfill
\begin{minipage}{0.45\linewidth}
\centering
\includegraphics[width=1.0\linewidth]{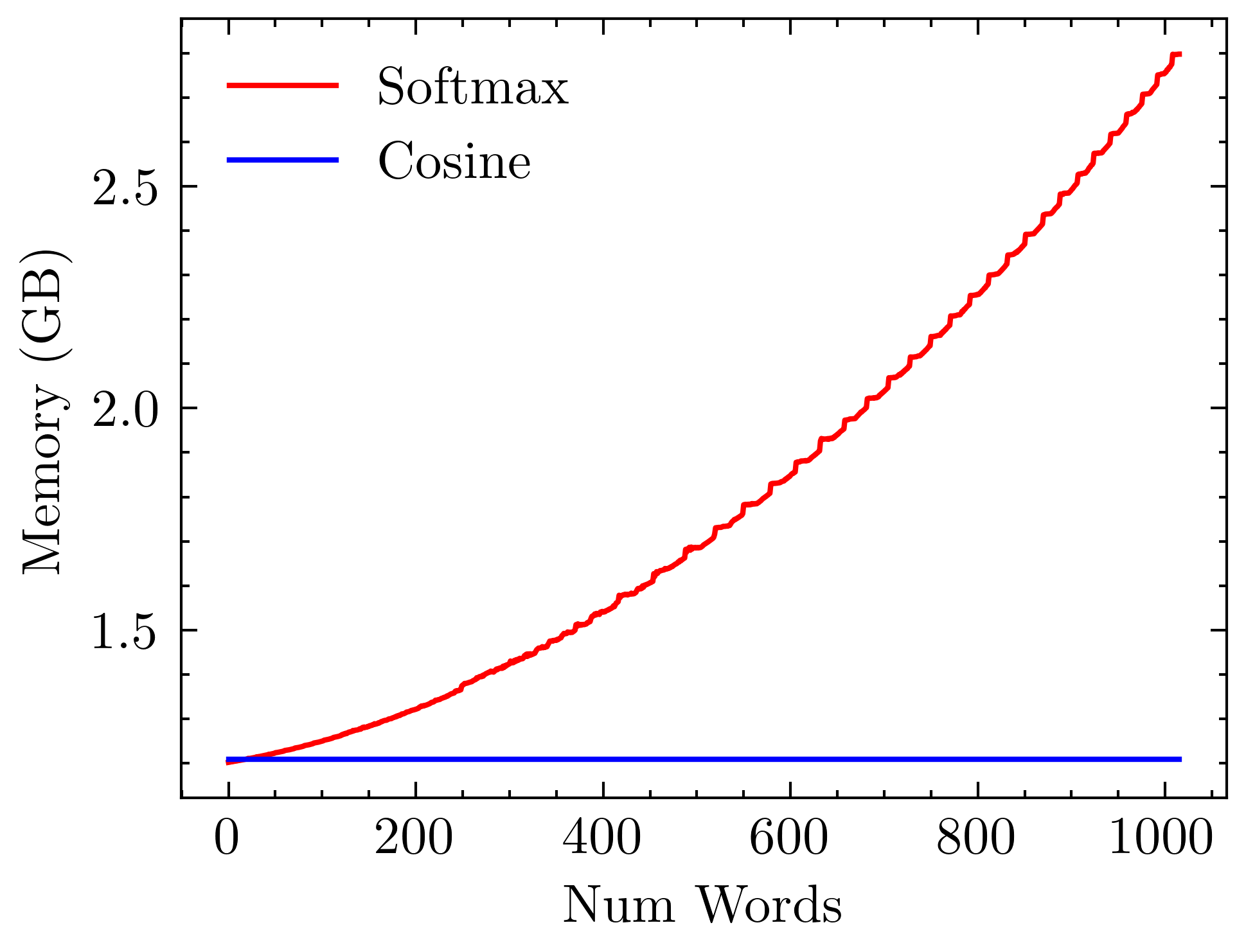}
\caption{Cosine attention has constant memory during inference while softmax attention has a quadratic increase under the naive implementation and linear increase using KV cache. This makes cosine attention more suitable for processing long sequences, especially in scenarios where memory is limited or the sequence length is not known in advance.}
\end{minipage}
\label{fig:RNN}
\end{figure*}

\subsection{Constant Memory During Inference}

One of the standout features of RNNs is their finite fixed-size hidden state. While this hidden state might restrict the amount of information it can store, it has the benefit of needing only a constant amount of memory during inference. This means cosine attention can perform inference with a constant memory footprint by storing the hidden state that results from the key-value outer product.

This property of cosine attention is particularly advantageous when compared to softmax attention. With softmax attention, the keys and values for all timesteps need to be cached during inference, which leads to a growing memory footprint as the context size increases. In contrast, cosine attention can simply accumulate all the keys and values into its hidden state, maintaining the same memory footprint regardless of the number of timesteps. 

Further, if the positional encodings are set up correctly, cosine attention can theoretically sample indefinitely without encountering any memory limitations due to the finite hidden state. This characteristic could make cosine attention an appealing choice for tasks that require processing long sequences or handling large context sizes. We leave the exploration of this potential behavior to future work, which we will discuss in section \ref{sec:future_work}.




\section{Results}

To assess the effectiveness of cosine attention, we conduct experiments on both bidirectional and causal attention scenarios. For the bidirectional case, we evaluate cosine attention using the BERT model, while for the causal case, we employ a variant of GPT-J. In our experiments, we train models using standard softmax attention and then replace the attention mechanism with cosine attention (Cottention), keeping all other architectural components unchanged. Both models were trained on eight 80 GB A100 GPUs via an NVIDIA DGX SuperPOD for approximately 5 days for BERT and 13 days for GPT-J. Although we focus on BERT for bidirectional tests, cosine attention can be applied to other models that utilize bidirectional attention, such as sequence-to-sequence translation \cite{vaswani2017attention} and text-to-image models like Stable Diffusion \cite{rombach2022high}.



\paragraph{BERT} BERT \cite{devlin2018bert} is a transformer model that leverages bidirectional attention. It is pre-trained on a combination of Wikipedia \cite{wikidump} (CC BY-SA 3.0 License, GFDL License) and BookCorpus \cite{Zhu2015Aligning} (GNU GPL License) using masked language modeling (MLM) and next sentence prediction (NSP) objectives. After pre-training, BERT is fine-tuned on the GLUE benchmark \cite{wang2018glue} (CC BY 4.0 License, Etc), which encompasses a diverse set of natural language understanding tasks. As shown in Table \ref{tab:BERT}, the BERT model with cosine attention achieves comparable performance to its softmax attention counterpart across the GLUE tasks.


\begin{table}[t]
\centering
\caption{Results of various BERT models on the GLUE benchmark.}
\vspace{4px}
\small 
\setlength{\tabcolsep}{4pt} 
\begin{tabular}{l c c c c c c c c c}
\hline
\rule{0pt}{3ex}
Model & MNLI-(m/mm) & QQP & QNLI & SST-2 & CoLA & STS-B & MRPC & RTE & Average \\[1ex]
\hline
\rule{0pt}{3ex}
BERT\textsubscript{BASE} & 84.6/83.4 & 71.2 & 90.5 & 93.5 & 52.1 & 85.8 & 88.9 & 66.4 & 79.6 \\[1ex]
\hline
\rule{0pt}{3ex}
BERT\textsubscript{softmax} & 81.8/82.5 & 86.5 & 89.9 & 90.5 & 80.5 & 78.3 & 90.0 & 67.9 & 83.1 \\[1ex]
\hspace{0.26em}BERT\textsubscript{cosine} & 80.6/81.1 & 86.2 & 89.3 & 90.1 & 77.8 & 76.5 & 88.6 & 66.4 & 81.8 \\[1ex]
\hline
\end{tabular}
\label{tab:BERT}
\end{table}



\begin{figure}[ht]
\centering
\begin{minipage}{0.45\linewidth}
\centering
\includegraphics[width=1.0\linewidth]{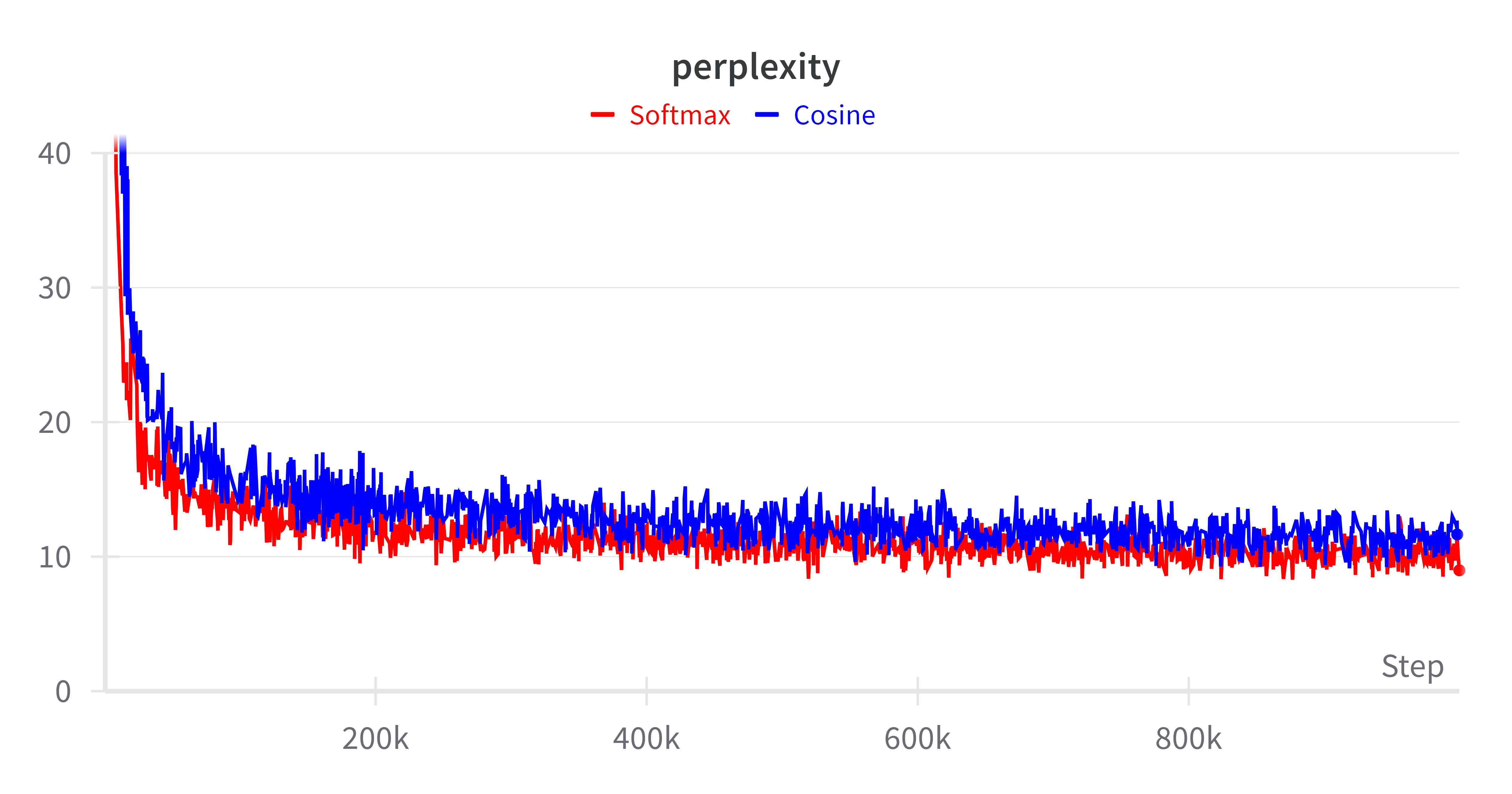}
\end{minipage}
\hfill
\begin{minipage}{0.45\linewidth}
\centering
\includegraphics[width=1.0\linewidth]{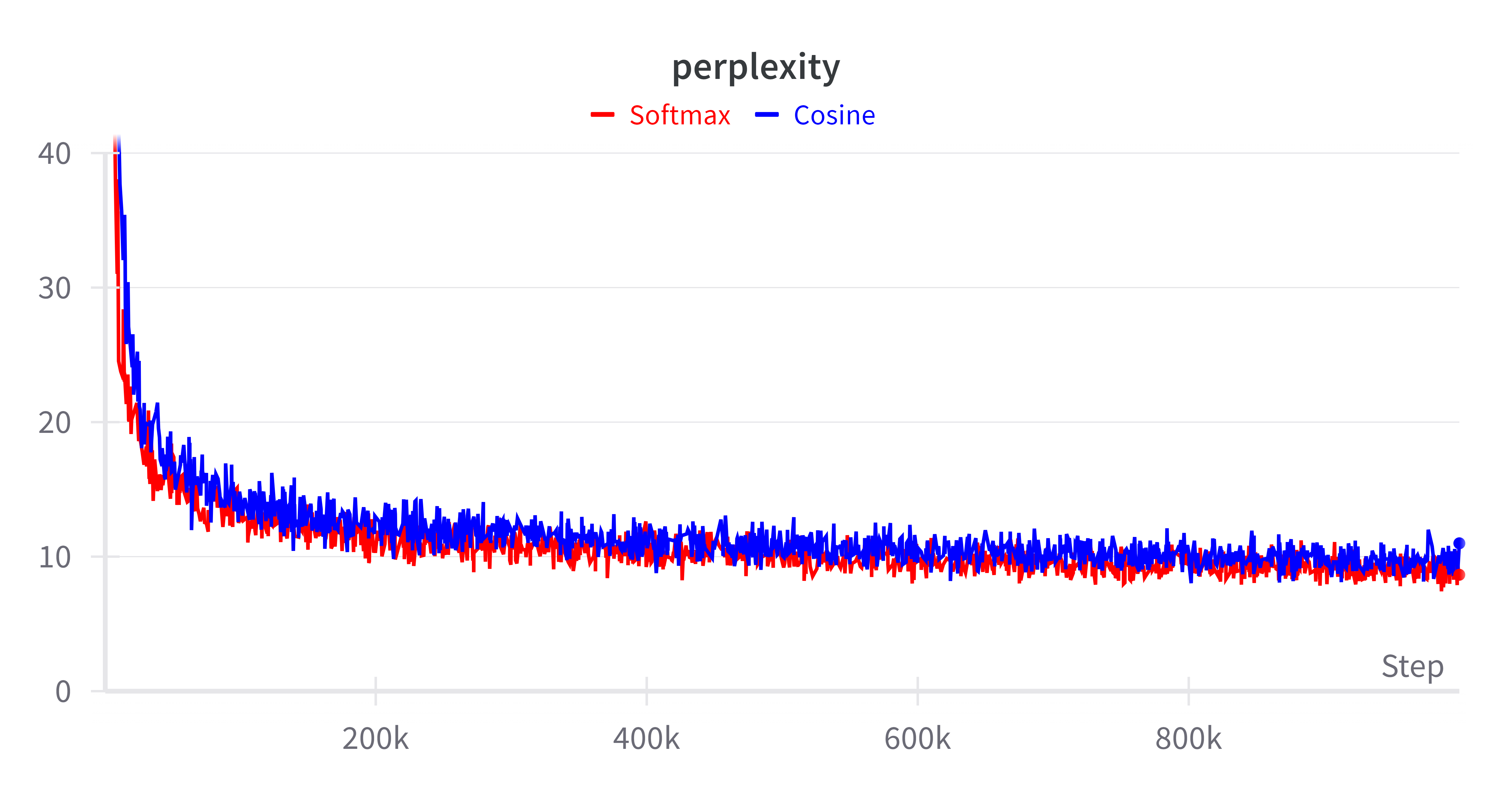}
\end{minipage}
\caption{Perplexity comparison for models with 300M (left) and 1.2B (right) parameters.}
\label{fig:GPT_loss}
\begin{minipage}{0.45\linewidth}
\centering
\includegraphics[width=1.0\linewidth]{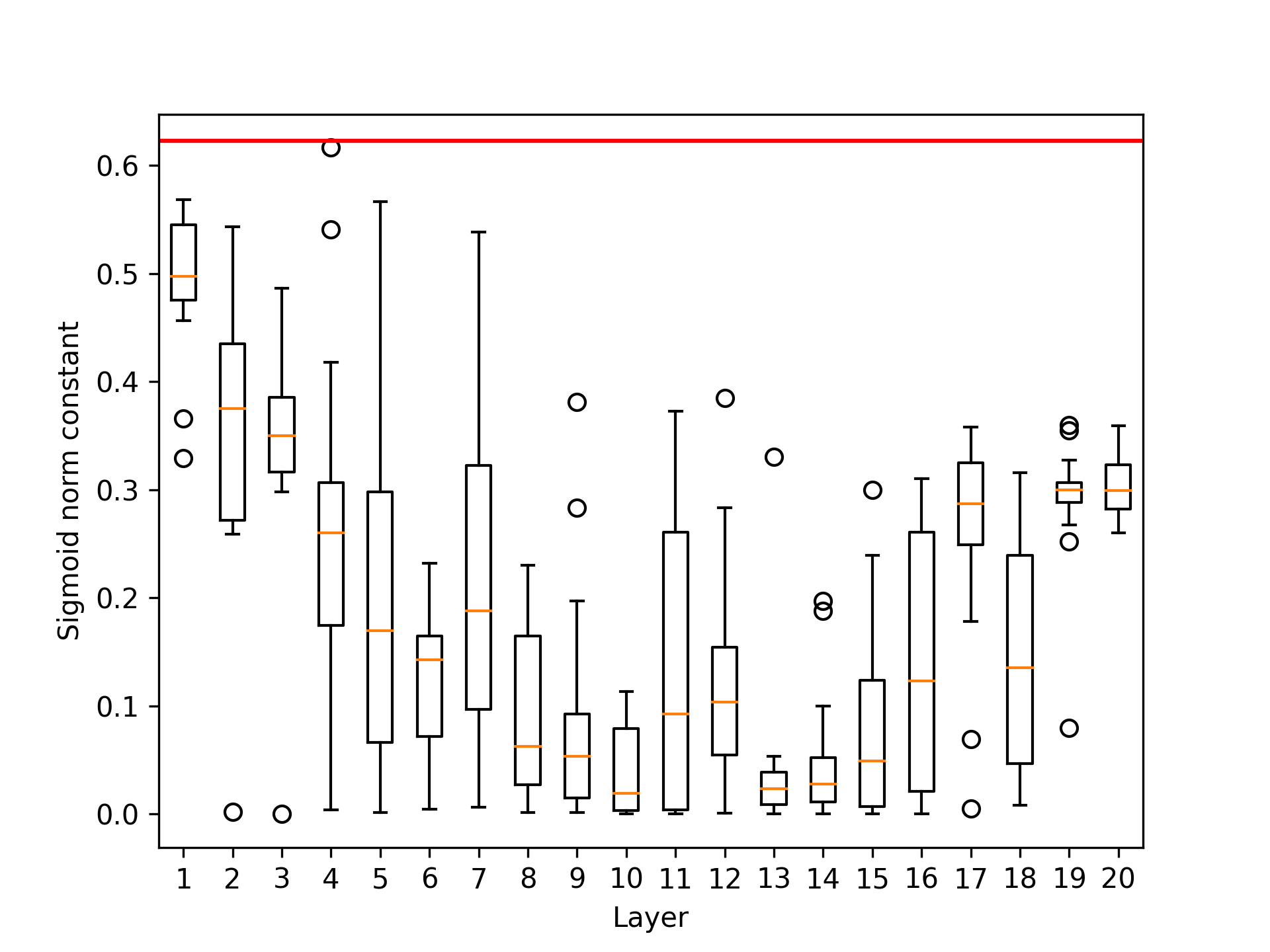}
\end{minipage}
\hfill
\begin{minipage}{0.45\linewidth}
\centering
\includegraphics[width=1.0\linewidth]{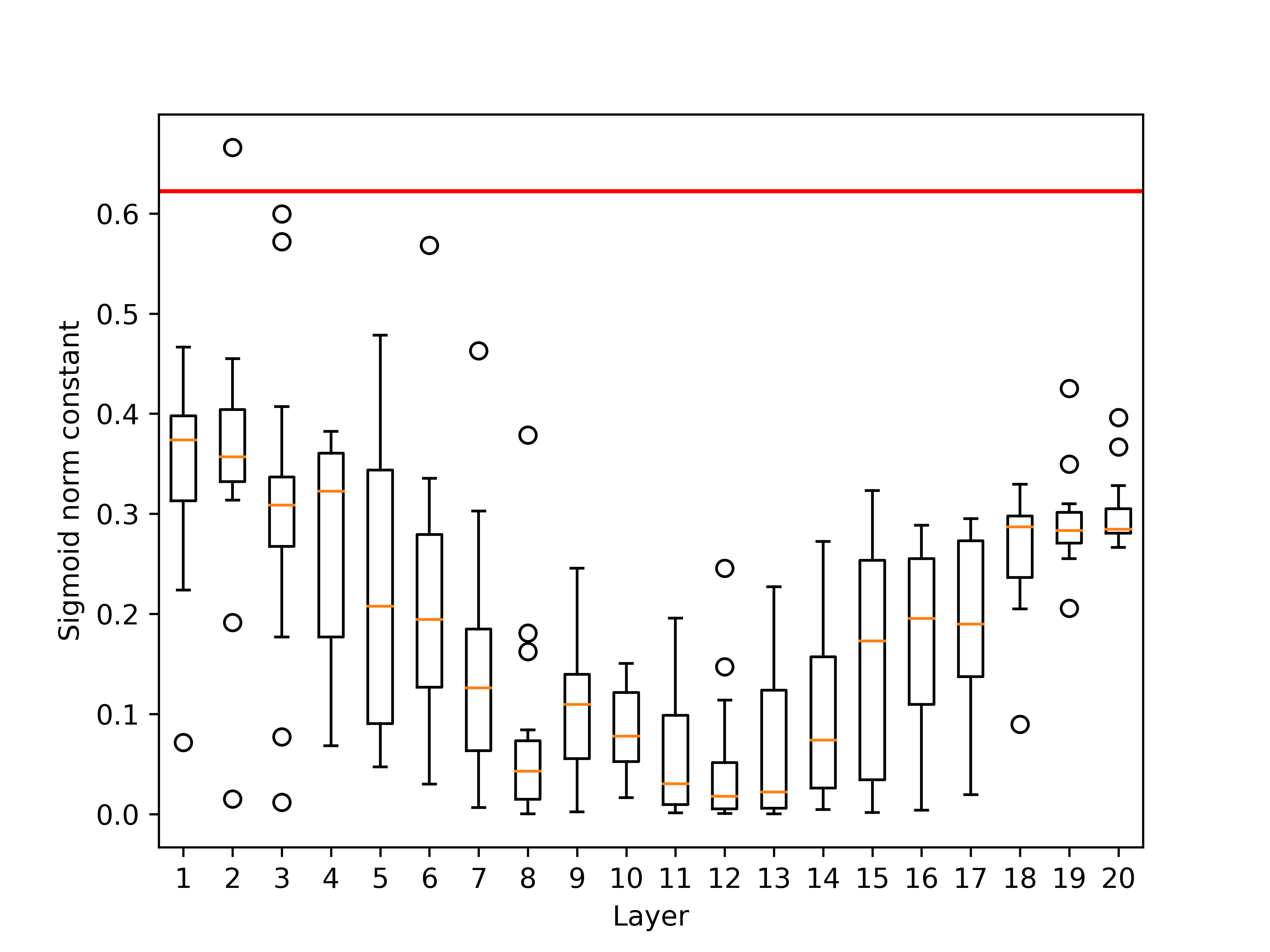}
\end{minipage}
\caption{Comparison of normalization constant distributions under sigmoid transform at end of training for the 300M parameter model (left) and the 1.2B parameter model (right). The red line denotes the initial value of all parameters.}
\label{fig:GPT_loss2}
\end{figure}

\paragraph{GPT} GPT-J \cite{wang2021gpt} is a causal transformer model trained on The Pile \cite{gao2020pile} (MIT License) using a next token prediction objective. The primary metric for evaluating such models is the loss or perplexity. We train two models with different parameter sizes: one with 300 million parameters and another with 1.2 billion parameters. Figure \ref{fig:GPT_loss} illustrates that GPT-J models with cosine attention achieve similar loss values compared to their softmax attention counterparts, demonstrating the effectiveness of cosine attention in causal language modeling tasks.

\subsection{Stabilization Constant}


An important observation from our experiments is the decay of the stabilization constant, $ m $, over the course of training. Initially set to $ 0.5 $, the value of this scalar parameter diminishes significantly by the end of the training process. We posit that the scalar plays a crucial role in stabilizing the model during the early stages of training when the randomly initialized parameters are in a highly unstable state. As training progresses and the model converges to a more stable configuration, the need for extra normalization gradually diminishes. Consequently, the model becomes less reliant on the stabilization constant, allowing it to adapt and learn more effectively from the data. This behavior suggests that the stabilization constant acts as a regularizer, helping to guide the model towards a stable and well-behaved solution space.

\begin{figure}[t]
\centering
\begin{minipage}{0.24\linewidth}
\centering
\includegraphics[width=\linewidth]{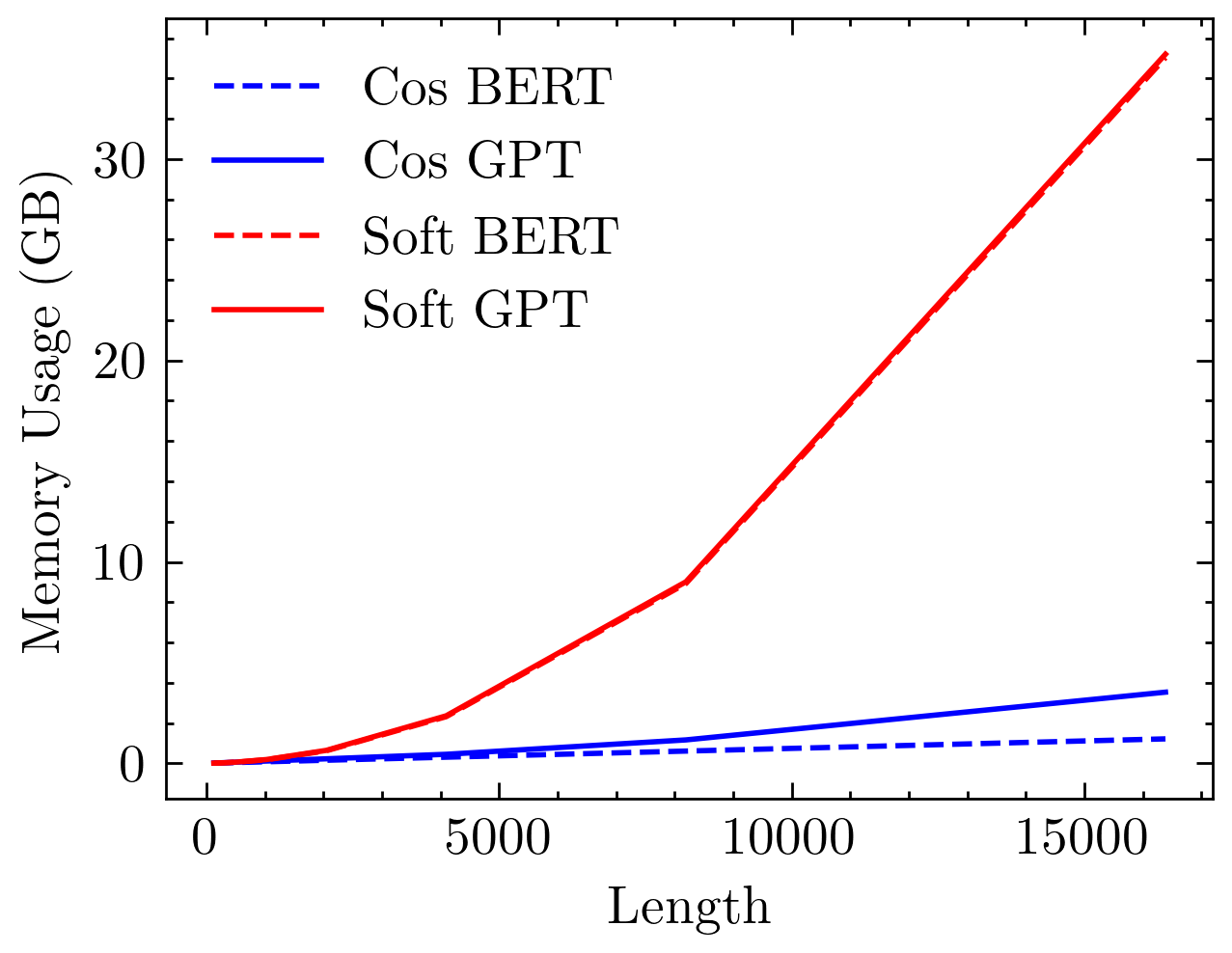}
\label{fig:len_mem}
\end{minipage}
\hfill
\begin{minipage}{0.24\linewidth}
\centering
\includegraphics[width=\linewidth]{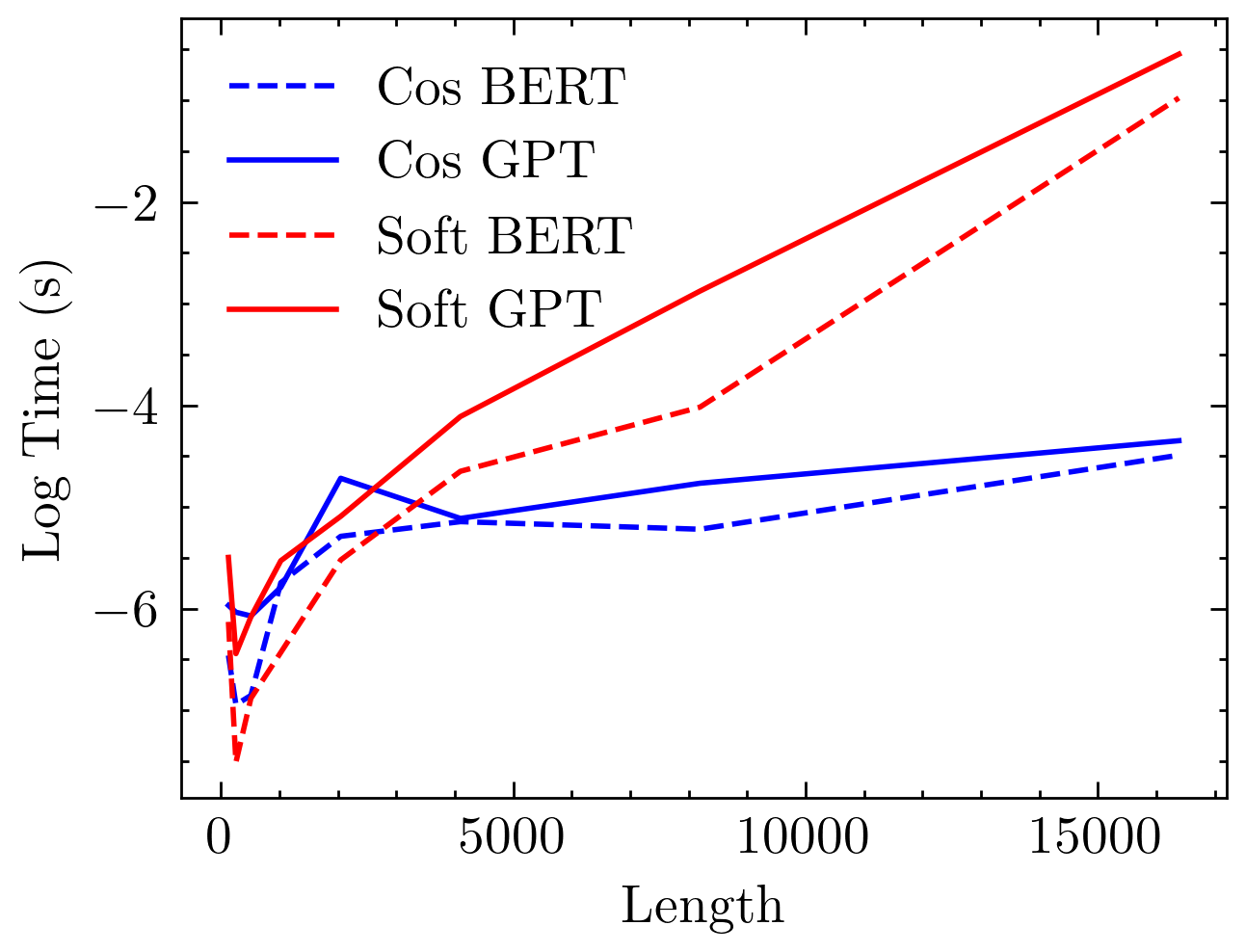}
\label{fig:len_time}
\end{minipage}
\hfill
\begin{minipage}{0.24\linewidth}
\centering
\includegraphics[width=\linewidth]{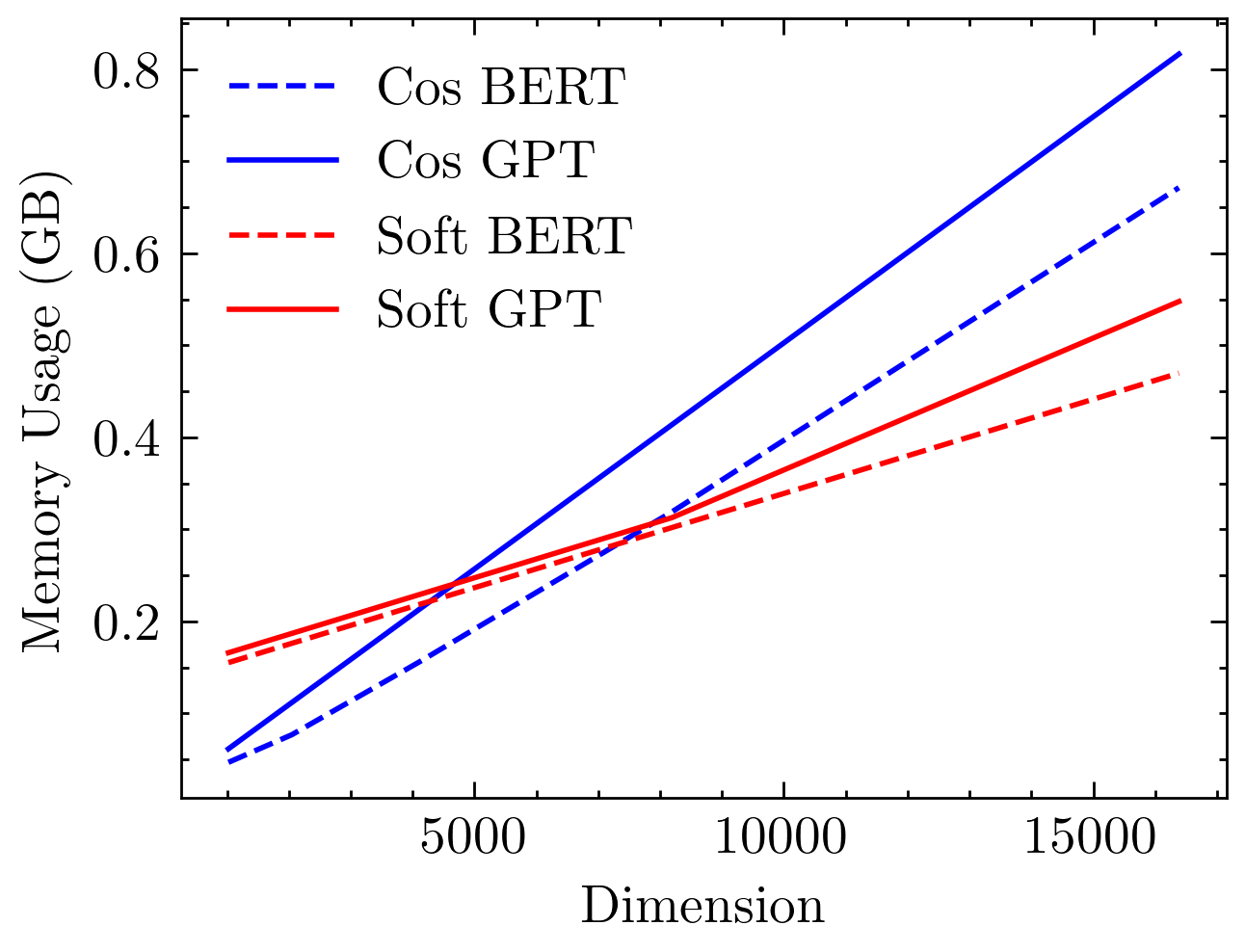}
\label{fig:dim_mem}
\end{minipage}
\hfill
\begin{minipage}{0.24\linewidth}
\centering
\includegraphics[width=\linewidth]{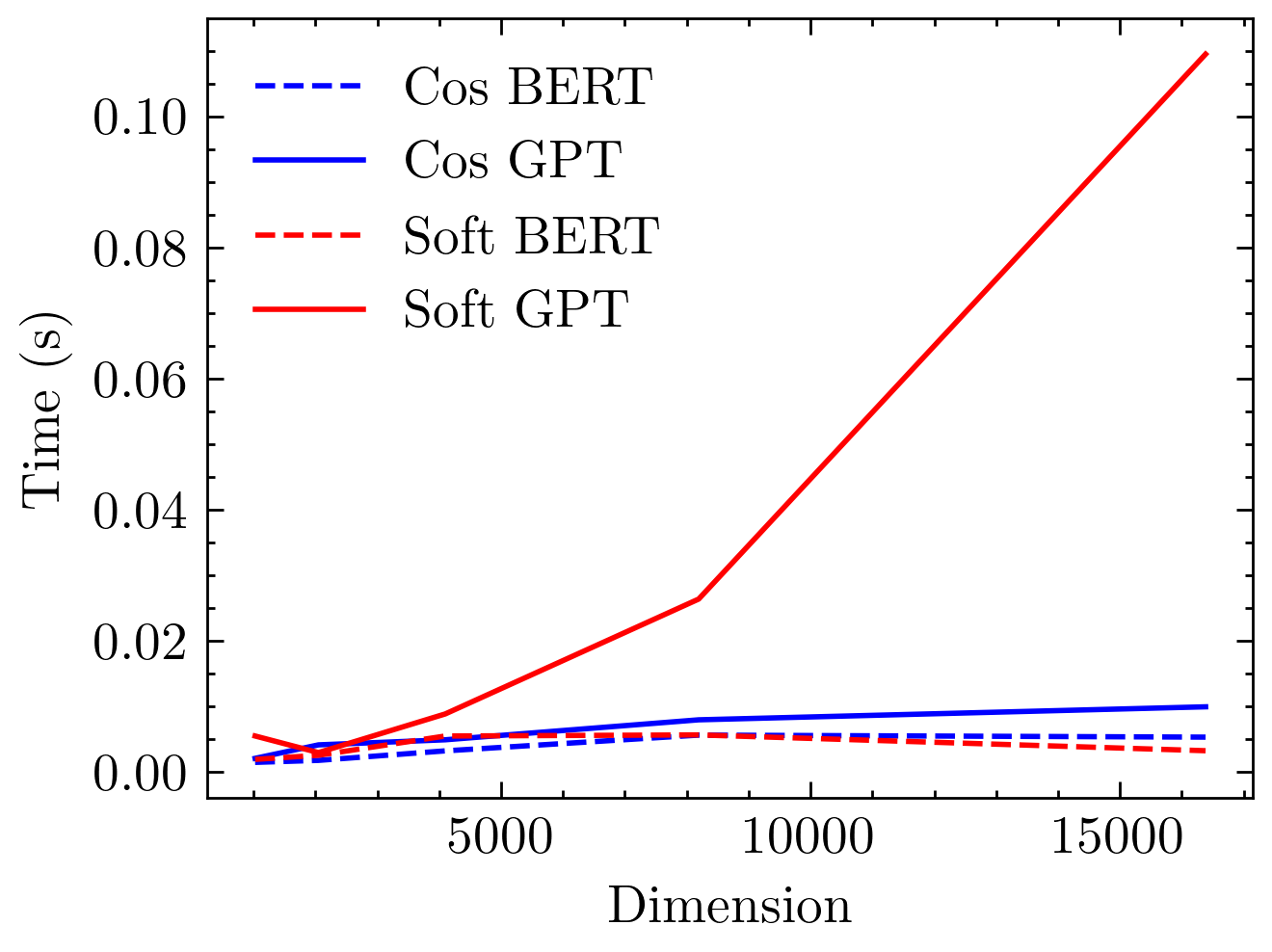}
\label{fig:dim_time}
\end{minipage}
\caption{Time and memory usage comparison between softmax and cosine attention models. Softmax models exhibit quadratic complexity, while cosine models demonstrate linear complexity with respect to sequence length. Interestingly, the memory usage of the cosine attention models doesn't seem to scale quadratically with respect to dimension.}
\label{fig:time_mem}
\end{figure}

\section{Performance Evaluation}

The theoretical memory complexity of cosine attention is expected to be linear with respect to the sequence length. We empirically validate this claim by examining the memory usage of both the bidirectional BERT and the causal GPT models, with the GPT model leveraging a custom CUDA kernel for efficient computation. Figure \ref{fig:time_mem} (top left) illustrates the linear relationship between memory usage and sequence length, confirming our theoretical expectations.

When considering the dimensionality of the input, cosine attention is expected to exhibit a quadratic increase in memory usage. However, our empirical findings, as shown in Figure \ref{fig:time_mem} (bottom left), reveal that the actual memory usage does not strictly follow a quadratic trend for either the bidirectional or causal case. This discrepancy between the theoretical and empirical results warrants further investigation to gain a deeper understanding of the factors influencing memory usage in relation to input dimensionality

In terms of time complexity, softmax attention has a complexity of $ \BigOSI{}{s^2d} $, while cosine attention has a complexity of $ \BigOSI{}{sd^2} $, where $ s $ denotes the sequence length and $ d $ denotes the dimensionality. The empirical time usage curves, depicted in Figure \ref{fig:time_mem} (top right, bottom right), exhibit patterns that closely resemble those observed for memory usage.

These empirical results highlight the potential advantages of cosine attention, particularly in terms of achieving linear memory complexity with respect to sequence length. This property makes cosine attention an attractive choice for processing longer sequences efficiently. However, the discrepancy between the theoretical and empirical memory usage trends regarding input dimensionality necessitates further exploration and analysis.

\section{Conclusions and Future Work}
\label{sec:future_work}

In this work, we introduced cosine attention, a novel attention mechanism that replaces the softmax function with cosine similarity. Our experimental results demonstrate that cosine attention achieves comparable performance to softmax attention on various natural language processing tasks while offering the benefits of subquadratic memory complexity and constant memory footprint during inference. Despite the success of cosine attention as a replacement for softmax, there remain challenges and opportunities for future work, described more fully below.

\paragraph{CUDA Kernel Optimization} The current implementation of the CUDA kernel for cosine attention is relatively basic and leaves room for optimization. By refining the algorithm and exploring advanced techniques, we aim to further improve the speed of computation, potentially surpassing that of softmax attention. This will involve investigating efficient parallelization strategies, memory access patterns, and kernel launch configurations.

\paragraph{Scaling Cosine Attention} In this work, we have applied cosine attention to relatively smaller models (BERT and GPT-J). However, the application of cosine attention to larger, state-of-the-art models was not investigated. Future research will focus on integrating cosine attention into these larger architectures and studying the effects on performance, scalability, and computational efficiency. This will provide valuable insights into the practicality and benefits of cosine attention.

\paragraph{Exploring Normalization Techniques} The normalization value $ m $ was set to $ 0.5 $ in our experiments, but this choice was not extensively tuned. Future work will involve a comprehensive analysis of different normalization techniques and their impact on model stability and performance. Additionally, while dividing by a power of the sequence length has shown to stabilize training, alternative approaches may yield better results. A thorough stability analysis will be conducted to identify the optimal normalization strategy for cosine attention.

\paragraph{Investigating Matrix Factorization Opportunities} One of the limitations of softmax attention is the constraint imposed by the $ QK^T $ multiplication, which restricts the exploration of alternative configurations. Cosine attention, on the other hand, decouples the computation of $ Q $, $ K $, and $ V $, opening up new possibilities for matrix factorization. Future research will delve into various factorization techniques that can take advantage of this decoupling, potentially leading to more efficient and effective attention mechanisms. Moreover, the decoupling of these matrices could enhance the performance of methods like LoRA and GaLORE by allowing them to operate on $ K^TV $ instead of $ QK^T $, potentially resulting in more efficient and expressive attention representations.

\paragraph{Leveraging RNN Formulation} The reformulation of cosine attention as a recurrent neural network (RNN) presents numerous opportunities for further optimization and analysis. Future work will explore techniques to leverage the RNN formulation for improved efficiency and performance. This may involve investigating advanced RNN architectures, such as long short-term memory (LSTM) or gated recurrent units (GRU), and adapting them to the cosine attention framework. Additionally, the RNN perspective may provide insights into the temporal dynamics of cosine attention and inspire novel approaches to capture long-range dependencies efficiently.

\paragraph{Addressing Limitations} While cosine attention has shown promising results, several limitations should be investigated through additional analysis. While our work showed the utility of cosine attention in several contexts, an analysis of resilience to input scale and instability was not systematically investigated. Comprehensive evaluations should be conducted on a wider range of tasks and datasets to assess the generalization and robustness of cosine attention across different domains. Through continued research and exploration, we hope to develop a more comprehensive understanding of cosine attention and its implications for various natural language processing tasks and beyond.

\medskip
\newpage
{
\small
\bibliographystyle{unsrt}
\bibliography{references}
}

\medskip
\newpage

\appendix

\section{Naive Cosine Attention Code}
\label{App_Code}

We provide a naive implementation of cosine attention using PyTorch. This implementation serves as a starting point for understanding the cosine attention mechanism and its computation process. However, it is not optimized for performance and does not utilize advanced techniques such as custom CUDA kernels. Further optimizations and improvements can be made to enhance the efficiency and scalability of the implementation. 

\vspace{.15in}
\begin{lstlisting}[language=Python]
import torch
    
class AttnMul(torch.autograd.Function):
    @staticmethod
    def forward(ctx, Q, K, V):
        ctx.save_for_backward(Q, K, V)
        return ((V.unsqueeze(-1) * K.unsqueeze(-2)).cumsum(-3)
            * Q.unsqueeze(-2)).sum(-1)

    @staticmethod
    def backward(ctx, grad_output):
        Q, K, V = ctx.saved_tensors
        grad_Q = ((V.unsqueeze(-1) * K.unsqueeze(-2)).cumsum(-3)
            * grad_output.unsqueeze(-1)).sum(-2)
        grad_K = ((grad_output.unsqueeze(-1) * Q.unsqueeze(-2))
            .flip(-3).cumsum(-3).flip(-3)
            * V.unsqueeze(-1)).sum(-2)
        grad_V = ((grad_output.unsqueeze(-1) * Q.unsqueeze(-2))
            .flip(-3).cumsum(-3).flip(-3)
            * K.unsqueeze(-2)).sum(-1)
        return grad_Q, grad_K, grad_V

def CosineAttention(Q, K, V, s, norm_const):
    # Q, K of shape (N, H, s, d_key)
    # V of shape (N, H, s, d_value)
    # s of shape (N, 1, 1, 1) is a scalar representing the sequence length at the current timestep.
    # norm_const of shape (1, H, 1, 1)
    Q = torch.nn.functional.normalize(Q, dim=-1, p=2)
    K = torch.nn.functional.normalize(K, dim=-1, p=2)
    V = V / s**norm_const.sigmoid()
    return AttnMul.apply(Q, K, V)
\end{lstlisting}

\end{document}